\documentclass[journal]{IEEEtran}

\usepackage{amsmath}
\usepackage{arydshln}
\usepackage{multirow} 
\usepackage{caption}
\usepackage{booktabs}
\usepackage{graphicx}
\usepackage{enumitem}
\usepackage[table]{xcolor}
\usepackage{colortbl}
\usepackage{multirow}
\usepackage{orcidlink}
\usepackage{array}
\usepackage{booktabs}
\usepackage{hyperref}
\hypersetup{
    colorlinks=false,
    pdfborder={0 0 0}
}
\definecolor{pi}{RGB}{230, 240, 255}   
\definecolor{pp}{RGB}{255, 245, 230}   
\definecolor{nu}{RGB}{240, 240, 240}   

\usepackage{amsmath,amssymb,amsfonts}
\usepackage{textcomp}
\usepackage{xcolor}
\usepackage{tikz}
\usepackage{subcaption}
\usepackage{float}
\usepackage{csquotes}
\usepackage{siunitx}
\usepackage{tabularx}
\usepackage{threeparttable}
\usepackage{multicol}
\usepackage{xcolor,colortbl}
\usepackage{caption}

\begin{document}
%
\title{Evolution of Multimodal Question Answering over Text, Tables, and Images: From Modality-Adaptive Extraction to Unified Language Representation}
%
%
%

\author{Abdullah Al Shafi \\ Department of CSE, Khulna University of Engineering \& Technology, Bangladesh \\ Email: \href{mailto:abdullah@iict.kuet.ac.bd}{abdullah@iict.kuet.ac.bd}
}

\maketitle
\thispagestyle{empty}

\begin{abstract}
The rapid growth of multimodal data has intensified the need for question answering (QA) systems capable of reasoning across heterogeneous sources such as text, tables, and images. In this paper, we present a comprehensive methodological comparison of three influential frameworks, namely Multimodal Adaptive Extraction (MAE), Solar, and UniMMQA, tracing the evolution of multimodal question answering from modality-adaptive pipelines to fully unified architectures. We examine how each approach models cross-modal interactions, transforms heterogeneous inputs, and performs reasoning, highlighting key design differences in modality representation, reasoning, and answer generation. Our analysis demonstrates a clear shift from explicit modality-specific processing toward unified text-centric formulations enabled by pre-trained language models (PLMs). Empirical comparisons across benchmark datasets show that this transition leads to substantial improvements in both Exact Match (EM) and F1-Scores, with UniMMQA achieving the most consistent and scalable performance. Despite these advances, we identify persistent challenges, including information loss during modality transformation, error propagation in multi-stage pipelines, and limitations in capturing fine-grained cross-modal dependencies. Overall, this study provides a deeper understanding of current design trends and offers insights into the future direction of unified multimodal reasoning systems.
\end{abstract}


%

\section{Introduction}
%
%
%
%
The rapid growth of digital information has led to an overwhelming information explosion, making it increasingly difficult for users to obtain concise and accurate answers \cite{abdel2023deep}. Question answering (QA) systems have therefore become an important solution to information overload, with early efforts focusing on knowledge-based, document-based, and community-based settings \cite{shao2023prompting, li2018unified}. However, these systems are typically limited to unimodal or single-source inputs, which restricts their applicability in real-world scenarios where relevant information is distributed across heterogeneous modalities.

To address this limitation, recent research has moved toward multimodal question answering (MMQA) and its conversational extension, multimodal conversational question answering (MMCoQA), where systems are required to reason over multiple modalities such as text, tables, and images within multi-turn dialogues \cite{hannan2020manymodalqa, talmor2021multimodalqa}. In this setting, a key challenge is that different questions may require different modalities, and effective reasoning often depends on jointly leveraging complementary information across sources.

In this work, we analyze three representative recent approaches that tackle multimodal and conversational QA over text, tables, and images, namely modality-adaptive extraction (MAE) \cite{li2022mmcoqa}, unified language representation with retrieval-augmented reasoning (Solar) \cite{yu2023unified}, unified language Representation with rationale-enhanced reasoning (UniMMQA) \cite{luo2023unifying}. Although these works share the same goal of enabling QA over heterogeneous modalities, they differ in how multimodal information is represented, retrieved, and reasoned. \cite{talmor2021multimodalqa} introduces the first ever conversational multimodal QA setting namely MMCoQA where questions evolve across dialogue turns and require reasoning over text, tables, and images. It highlights the difficulty of maintaining contextual understanding while selecting relevant evidence across modalities in a conversational setting. Then, Yu et al. \cite{yu2023unified} proposes a unified language representation framework called Solar that converts multimodal inputs into a text-centric representation, enabling a single QA model to handle heterogeneous sources. This approach simplifies multimodal reasoning by leveraging large pre-trained language models (PLMs), while relying on retrieval and ranking over unified textual evidence.  After that, the authors in \cite{luo2023unifying} further improves modality unification by transforming text, tables, and images into a shared textual space with rationale, enabling end-to-end multimodal QA using text-to-text generation models (T5). This line of work emphasizes simplicity and scalability by reducing multimodal QA to a unified generation problem.

Despite their differences, all three approaches reflect a common trend, shifting from modality-specific modeling toward unified text-centric multimodal reasoning, driven by the strong generalization capability of PLMs. However, challenges remain in preserving fine-grained multimodal information, maintaining cross-modal alignment, and handling complex conversational dependencies.

\begin{figure*}
\centerline{\includegraphics[width=0.99\textwidth]{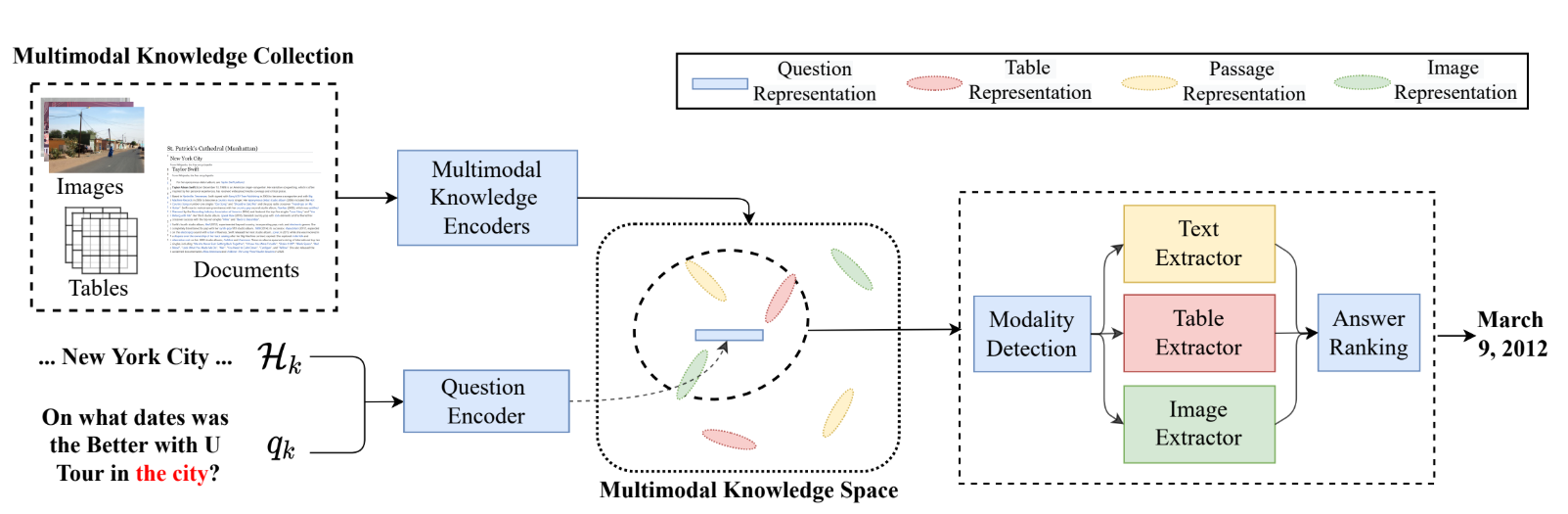}}
\caption{Visualization of Multimodal Adaptive Extraction (MAE) Pipeline \cite{li2022mmcoqa}.}
\label{fig:mae}
\end{figure*}


This paper provides a structured review of recent progress in multimodal and conversational question answering over heterogeneous sources. The main contributions are:

\begin{itemize}
    \item We present a systematic comparison of three representative MMQA/MMCoQA frameworks, focusing on how they handle text, table, and image modalities within QA systems.
    \item We analyze the evolution toward unified multimodal representations, highlighting how recent methods increasingly reformulate multimodal QA as a text-centric problem using pre-trained language models.
    \item We identify the key limitations of existing approaches, including information loss during modality transformation, limited cross-modal alignment, and challenges in conversational reasoning across heterogeneous sources.
\end{itemize}

The remainder of the paper is organized as follows. Section \ref{sec:method} shows the methodological evoluation in the field of MMQA. Sections \ref{sec:setup} and \ref{sec:result} describe the experimental setup and results respectively. Finally, Section \ref{sec:conclusion} concludes the paper.

\section{Literature Review}
\label{sec:literature}
The research field of multimodal question answering develops through three main research areas which include conversational question answering, multimodal reasoning across different types of sources, and multimodal large language models (MLLMs) .


The research field of conversational QA established its early benchmarks through the development of CoQA \cite{reddy2019coqa} and QuAC \cite{choi2018quac} which demonstrated the necessity of dialectical context modeling. The latest research seeks to enhance contextual comprehension and system durability through the application of PLMs together with retrieval-augmented methods \cite{lewis2020retrieval}. However, most current conversational QA systems still restricts their functions to handling text input, thus they fail to support multimodal reasoning capabilities.


The development of MMQA progressed through the creation of datasets which require simultaneous reasoning in textual content, tabular data and visual elements. MultimodalQA \cite{talmor2021multimodalqa} and ManyModalQA \cite{hannan2020manymodalqa} demonstrate the necessity of performing compositional reasoning to access multiple types of information. Recent research shows a growing trend toward using unified system designs together with extensive pre-trained models to enhance cross-modal reasoning capabilities and system scalability. The process of connecting different modalities and managing various data distribution patterns presents ongoing difficulties.


Vision-language pretraining has made significant advancements through the development of large-scale models which learn from image-text pairs. The systems BLIP-2 \cite{li2023blip} and PaLI \cite{chen2022pali} utilize pretrained vision encoders together with extensive language models to achieve high-performance results under zero-shot and few-shot testing conditions. The research shows that increasing the size of multimodal pretraining programs results in better outcomes for various tasks. However, the systems focus on perception and generation tasks but they lack optimization for intricate multi-hop reasoning tasks.


More recently, the emergence of MLLMs has reshaped the field. The various models which include Kosmos-1 \cite{huang2023language}, Kosmos-2 \cite{peng2023kosmos}, LLaVA \cite{liu2023visual}, and MiniGPT-4 \cite{zhu2023minigpt} use instruction tuning and cross-modal alignment to enable large language models to understand visual information. The systems can process multiple types of input simultaneously while executing different reasoning tasks within one system. Nevertheless, they often exhibit limitations in factual grounding and may produce hallucinated outputs when reasoning over complex multimodal contexts.


Another line of research focuses on grounded and structured multimodal reasoning, aiming to improve the reliability and interpretability of model predictions. Researchers in recent studies investigate how to combine structured data representations with retrieval systems to improve reasoning performance in multimodal systems according to sources \cite{liu2025retrieval, cui2024more, lee2024multimodal}. The methods boost task performance through their specific capabilities, but they need extra supervision and special system designs, which restrict their application to other situations.



\section{Methodological Evolution}
\label{sec:method}
Recent multimodal QA approaches have evolved from modality-adaptive architectures such as MAE \cite{li2022mmcoqa} to unified language-based frameworks like Solar \cite{yu2023unified}, and further toward end-to-end reasoning-enhanced models such as UniMMQA \cite{luo2023unifying}.

\begin{figure*}
\centerline{\includegraphics[width=0.99\textwidth]{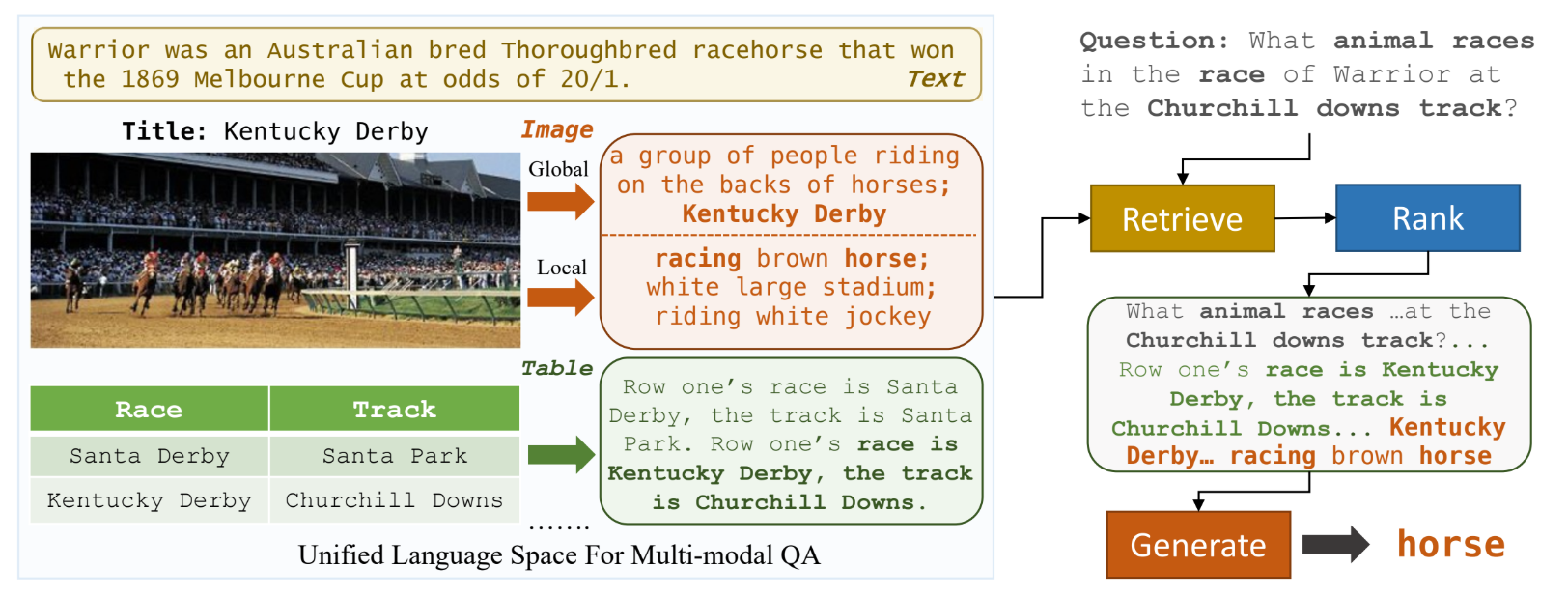}}
\caption{Visualization of Solar Pipeline \cite{yu2023unified}.}
\label{fig:solar}
\end{figure*}

\subsection{Multimodal Adaptive Extraction: MAE \cite{li2022mmcoqa}}


To address the challenges of MMCoQA, the Multimodal Conversational Question Answering system with Adaptive Extractors (MAE) \cite{li2022mmcoqa} introduces a baseline framework that explicitly models modality selection, contextual understanding, and answer extraction across heterogeneous sources. The system is designed to operate over a collection of multimodal evidences, including textual passages, tables, and images, while incorporating conversational history to resolve context-dependent queries.

Let the current question at turn $k$ be denoted as $q_k$, and the corresponding conversational history be $H_k = \{q_1, a_1, \dots, q_{k-1}, a_{k-1}\}$. The system is also given a multimodal knowledge collection $C = \{C_p \cup C_t \cup C_i\}$, where $C_p$, $C_t$, and $C_i$ represent passages, tables, and images, respectively. The objective is to retrieve relevant evidence from $C$ and generate the answer $\hat{a}_k$ for the current query.

At a high level, the MAE framework decomposes the task into three sequential components as shown in Fig. \ref{fig:mae}: (i) conversational question understanding, (ii) multimodal evidence retrieval, and (iii) adaptive answer extraction.

\textbf{Conversational Question Understanding:} First, the model incorporates conversational context by reformulating the current query using information from $H_k$, allowing it to resolve coreference and implicit references. The reformulated query $q'_k$ is then encoded into a dense vector representation using a transformer-based encoder, i.e., $v_q = W_q F_q(q'_k)$.

\textbf{Multimodal Evidence Retrieval:} Modality-specific encoders are used to transform different types of knowledge into a unified embedding space. For textual passages, standard transformer-based encoders are applied directly to obtain semantic representations. For tables, each table is first linearized into a sequence format using row-wise concatenation of headers and cell values and then processed as text input by the encoder. For images, visual features are extracted using a pretrained ResNet, which converts each image into a fixed-dimensional feature vector. These representations are projected into the same embedding space and obtained semantic representation $v_p, v_t, v_q$ for knowledge collection of textual passages, tables, and images respectively.

After that, the similarity scores $s_a$ between the question representation $v_q$ and all candidate knowledge items is calculated, and selects the top-$N_r$ most relevant items as evidence. These retrieved items may originate from different modalities, providing diverse supporting information for answering the question.

\textbf{Adaptive Answer Extraction:} A key aspect of this design is the adaptive selection process, which determines the most suitable extractor based on the input query and retrieved evidences. The model first predicts the most suitable modality for the question as a multi-class classification task, producing a probability distribution over text, table, and image modalities. Based on this prediction with highest probability ($s_b$), a corresponding extractor is selected. The TextExtractor and TableExtractor operate as span prediction models that identify answer spans within the input sequences, while the ImageExtractor combines visual features with textual representations of candidate answers to infer the most probable answer with extraction probability $s_c$. Finally, Each candidate answer is assigned a score as a sum of retrieval relevance ($s_a$), modality confidence ($s_b$), and extraction probability ($s_c$), and the highest-scoring answer is returned as the final output.

Despite its modular design, MAE has several limitations. Although it retrieves evidence from multiple modalities, the model ultimately relies on a single predicted modality during the extraction stage. This design restricts its ability to handle questions that require joint reasoning across multiple modalities, where complementary information must be integrated. Furthermore, the reliance on modality prediction introduces potential errors, as incorrect classification may lead to the selection of an inappropriate extractor and degrade overall performance. These limitations highlight the need for more advanced approaches that enable unified cross-modal reasoning in question answering.

\begin{figure*}
\centerline{\includegraphics[width=0.99\textwidth]{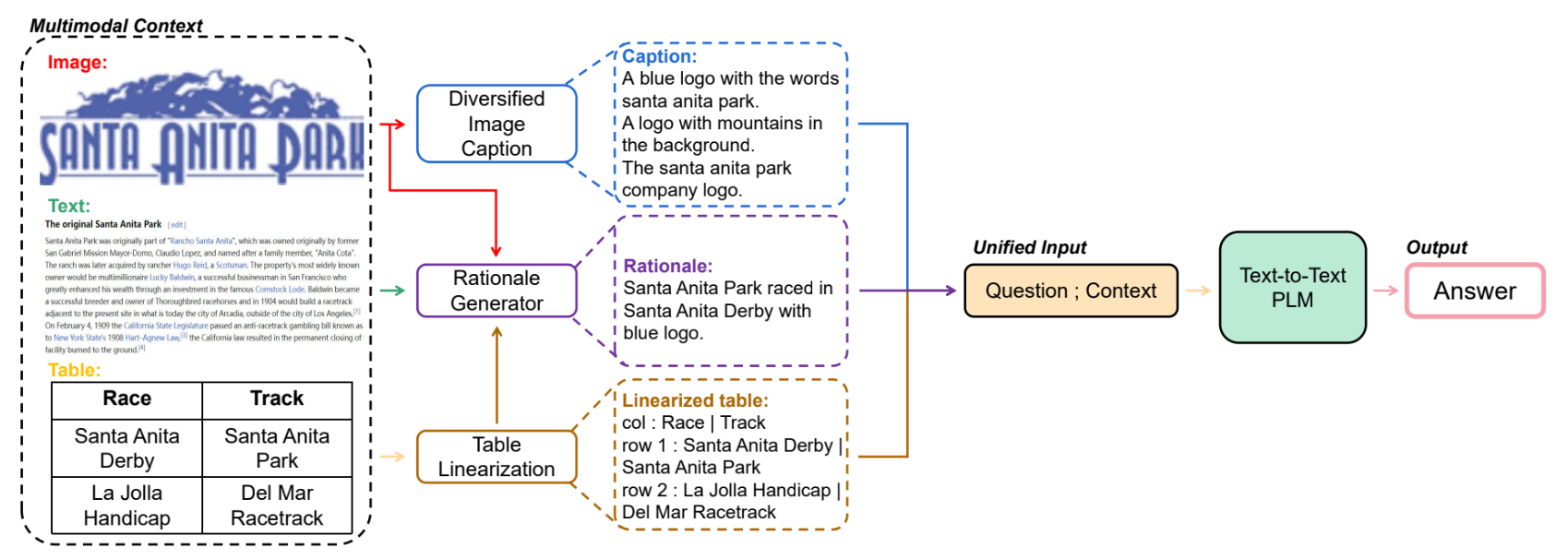}}
\caption{Visualization of UniMMQA Pipeline \cite{luo2023unifying}.}
\label{fig:unimmqa}
\end{figure*}

\subsection{Unified Language Representation with Retrieval-Augmented Reasoning: Solar \cite{yu2023unified}}
The Solar framework \cite{yu2023unified} presents multimodal question answering through transforming all modalities including text, tables, and images into one unified linguistic representation, followed by a unified QA pipeline operating solely over text.



\textbf{Unified Language Representation (Modality Conversion Layer):} The first core component of Solar transforms all input modalities into sequences of natural language, which enables modality-agnostic downstream processing.


Solar uses natural language templates to create deterministic linearization transformation for tabular data. The table data transforms into complete sentences through the process of converting each column and its respective value into spoken language. For example, from Fig. \ref{fig:solar} the first row is converted into \textquote*{Row one’s race is Santa Derby, the track is Santa Park}.


For images, the transformation is inherently lossy due to the absence of a native linguistic structure. The dual textualization system of Solar using both global and local representation provides mitigation of this problem. The global representation uses an image captioning model to create a complete scene description which includes necessary background information from the image title. The global captions provide an incomplete view of the information needed for QA because they miss essential detailed content. The local representation contains semantic elements that the object–attribute detection model has extracted through its detection of object types and attributes and actions. The local descriptors function as evidence at the micro-level which supports the main argument. The system creates an image representation by joining all global and local textual components through punctuation marks which results in a complete multi-level textual representation of the original image. The dual strategy preserves both complete contextual information and specific details which support reasoning processes.


For questions, especially in conversational settings, Solar does not treat each query independently. Instead, it constructs a contextualized query representation by concatenating the current question with prior dialogue history (previous questions and answers). This enables the model to implicitly capture co-reference, ellipsis, and conversational dependencies within a purely textual formulation.


\textbf{Unified Multimodal QA Model:} The process of mapping different modalities into the same textual space Z enables Solar to redefine multimodal QA as a retrieval-augmented text generation (RAG) problem. The pipeline consists of three sequential stages:

\begin{enumerate}
    \item \textbf{Retrieval:} The system first identifies a subset of relevant textual evidence from the unified space. The system uses dense retrieval mechanism to encode both the query and textual evidence into a shared embedding space through BERT. The system calculates similarity between the query and candidate clues through inner product computation which results in selection of top-K most relevant clues. The process establishes evidence from different sources to support the query which enables effective search space reduction.
    \item \textbf{Ranking:} The retrieved candidates may still contain noise or weakly relevant information. Solar introduces its cross-encoder ranking model which simultaneously evaluates query input together with each candidate clue. The ranking model uses cross-attention to process the combined sequence of query and clue elements which enables detailed token interactions. Each candidate is assigned a relevance score via a sigmoid classifier. The top-N ranked clues are retained for the final stage.
    \item \textbf{Generation:} The final answer is produced using a sequence-to-sequence generator (e.g., T5). The model receives its input through the combination of the query and the highest-ranked clues. The different types of clues now appear in the same format which allows for uninterrupted logical deductions between them.

    
\end{enumerate}

However, the framework requires high-quality modality-to-text conversion because image captioning and object detection errors lead to prediction errors which affect the entire pipeline. The process of image textualization creates inherent data loss because vital visual elements including spatial relationships and subtle visual elements cannot be represented in textual format. Furthermore, the process of transforming structured data into textual format results in a loss of clear structural connections that exist in the original data. The multi-stage retrieve–rank–generate pipeline creates error accumulation through its design because it lacks ability to retrieve missing evidence during the retrieval process.


\subsection{Unified Language Representation with Explicit Rationale-enhanced Reasoning: UniMMQA \cite{luo2023unifying}}






The UniMMQA framework \cite{luo2023unifying} is designed to solve MMQA by transforming heterogeneous inputs, including text, tables, and images into a unified textual representation and then leveraging a text-to-text PLMs for answer generation. Given a question $Q$ and a multimodal context $C = \{I, T, P\}$, where $I$, $T$, and $P$ denote image, table, and passage respectively, the framework systematically converts each modality into text while preserving as much structural and semantic information as possible. As illustrated in the \ref{fig:unimmqa}, all transformed components are concatenated into a unified sequence and processed by a generative PLM (T5) to produce the final answer.

\textbf{Diversified Image-to-Text Transformation:} This is a core component of UniMMQA, which goes beyond conventional single-caption generation. UniMMQA first utilizes OCR technology to extract all visible textual components from images which include hidden texts that are found on signs and labels. Then, the system implements a vision-language captioning model which uses BLIP technology through a decoding method that combines Top-$K$ \cite{fan2018hierarchical} and Top-$p$ \cite{holtzman2019curious} sampling techniques instead of taking greedy one. The system generates each subsequent token through a probability distribution $P(I^{text}_t \mid I^{text}_{1:t-1})$ which first selects from the most probable tokens through Top-$K$ and then enables multiple token selection through Top-$p$ for balanced output. The system generates multiple captions which work together to show detailed visual information through their respective captions which display specific objects and background details that traditional systems would not display. The textual content of the captions together with the OCR results creates an enhanced written representation of visual material.

\textbf{Table-to-Text Conversion:} The UniMMQA system does not use template-based summarization because this method fails to maintain essential attributes needed for the task. The system uses a position-enhanced linearization method which creates a representation of content together with structural elements. The table serialization process organizes data in row format while using distinct indicators to define table elements. The table starts with headers marked as ``header :'' followed by rows which use the indexing system (e.g., ``row 1 : \ldots'') to identify their position. The table with $M$ rows and $N$ columns transforms into a sequence which maintains all cell contents together with their original positions. The system uses explicit encoding to protect relationships between column alignment with row identity which enables the model to retrieve precise data about attributes including height through template-based methods which discard unselected columns.


\textbf{Multimodal Rationale Generation:} Beyond simple modality unification, UniMMQA introduces a multimodal rationale generator to explicitly model cross-modal interactions. The system creates natural language explanations as intermediate outputs which describe how different modalities connect with each other through their relationships. A vision encoder (CLIP) extracts visual features $I_{feature}$ from the system while the system uses a transformer-based language encoder to encode textual representations which include passages and linearized tables into $P_{feature}$.  The system combines these two representations which then proceed to the generative decoder for producing the rationale $R = F_r(I_{feature}, P_{feature})$, where $F_r$ is the rationale generation model. 


\textbf{Answer Generation:} The answer generation stage processes the complete problem as a sequence-to-sequence assignment. The question $Q$ together with the unified textual context $C^{text} = \{I^{text}, T^{text}, P, R\}$ creates a single input sequence that is passed to a pretrained model T5. The model uses negative log-likelihood as its fine-tuning objective to produce output tokens which match the ground truth answer sequence. 


Despite its effectiveness, UniMMQA suffers from several limitations. The conversion of images into text, even with diversified captioning, may still miss fine-grained visual details (e.g., exact object counts), leading to incorrect reasoning. Similarly, although position-enhanced table linearization preserves all cell information, it significantly increases input length and complexity, which can negatively affect model efficiency and reasoning performance. Furthermore, both the image captioning and rationale generation modules are prone to hallucination, potentially introducing inaccurate or fabricated information that misguides the final answer generation.

\section{Experimental Setup}
\label{sec:setup}



\subsection{Datasets}
Experiments were conducted on several public benchmark datasets, including ManymodalQA \cite{hannan2020manymodalqa}, MultimodalQA \cite{talmor2021multimodalqa}, and MMConvQA \cite{li2022mmcoqa}. ManymodalQA contains 10,190 questions, including 2,873 images, 3,789 text and 3,528 tables, which have been split into train/dev sets. MultimodalQA consists of 29,918 question-answer pairs cross image, table and text modalities, 35.7\% of which require cross-modal reasoning. Extending the MultimodalQA \cite{talmor2021multimodalqa} benchmark, MMConvQA contains 1,179 conversations and 5,753 question-answering pairs. There are 218,285 passages, 10,042 tables and 57,058 images in MMConvQA, and about 24.4\% of conversations require reasoning cross three modalities.


\subsection{Evaluation Metrics}
Exact Match (EM) and F1 were adopted as the evaluation metrics for all datasets. EM measures whether the predicted answer exactly matches the ground-truth answer after normalization, assigning a score of 1 for an exact match and 0 otherwise. In contrast, F1 evaluates the token-level overlap between the predicted and ground-truth answers by considering both precision and recall, thereby providing partial credit for answers that are not exact but share common tokens.

\subsection{Baseline Models}
Several baseline models were used for comparison, including open-retrieval conversational QA system (ORConvQA) \cite{qu2020open}, multimodalQA model (ManyModalQA) \cite{hannan2020manymodalqa}, Multi-Hop Implicit Decomposition (Implicit-Decomp) \cite{talmor2021multimodalqa}, MMQA-T5-Large \cite{yoran2022turning},  Pre-traind for Reasoning Model (PReasM) \cite{yoran2022turning}, Structured Knowledge and Unified Retrieval-Generation (SKURG) \cite{yang2023enhancing}, Multimodal Graph Transformer (MGT) \cite{he2023multimodal}.

ORConvQA \cite{qu2020open} addresses conversational question answering in an open-retrieval setting by integrating a trainable retriever, a reranker, and a reader all based on Transformer, enabling end-to-end optimization of the retrieval and answer extraction process. In contrast, ManymodalQA \cite{hannan2020manymodalqa} adopts a modality-aware pipeline, where a question-type classifier first identifies the relevant modality and then routes the query to the corresponding unimodal QA system.

Implicit-Decomp \cite{talmor2021multimodalqa} uses programming techniques to break down complex questions into multiple reasoning steps that it solves through different modalities. The system generates programs which execute various single-modal and multi-modal QA components to produce the ultimate response. MMQA-T5 system uses sequence-to-sequence modeling together with table linearization to train T5 for multimodal QA applications. The system uses Implicit-Decomp for image-based questions which lets it integrate generation-based and decomposition-based methods.


PReasM \cite{yoran2022turning} improves T5 through synthetic tabular data which enhances its reasoning abilities but maintains the existing routing system for processing image-based inquiries. SKURG \cite{yang2023enhancing} uses knowledge graphs to model connections between different modalities which enables more organized and linked reasoning processes. MGT \cite{he2023multimodal} combines graph-based models with Transformer systems to enable multimodal graph learning which analyzes interactions among diverse data types.

\begin{table}[ht]
\centering
\caption{Experimental results on MultimodalQA.}
\begin{tabular}{lcccc}
\toprule
\multirow{2}{*}{Method} & \multicolumn{2}{c}{Dev} & \multicolumn{2}{c}{Test} \\
\cmidrule(lr){2-3} \cmidrule(lr){4-5}
 & EM & F1 & EM & F1 \\
\midrule
Implicit-Decomp & 48.8 & 55.5 & 49.3 & 55.9 \\
MMQA-T5-Large & 57.9 & 64.3 & 57.0 & 63.4 \\
PReasM-Large & 59.0 & 65.5 & 58.3 & 64.6 \\
SKURG & 59.4 & 63.8 & -- & -- \\
MGT & 52.1 & 57.7 & -- & -- \\
\midrule
Solar & 59.8 & 66.1 & -- & -- \\
\midrule
UniMMQA (T5-Base) & 67.9 & 74.0 & 67.0 & 72.9 \\
UniMMQA (T5-Large) & 71.3 & 77.1 & 70.1 & 75.8 \\
UniMMQA (T5-3B) & \textbf{75.5} & \textbf{81.7} & \textbf{73.7} & \textbf{80.1} \\
\bottomrule
\end{tabular}
\label{tab:mmqa}
\end{table}

\section{Experimental Result}
\label{sec:result}
\subsection{Overall Performance}
\textbf{MultimodalQA:} Table \ref{tab:mmqa} presents a comparison of UniMMQA and Solar against several baseline methods on the MultimodalQA dataset. The baseline testing results show that Implicit-Decomp performs poorly because it can only process multimodal reasoning through its separate uni-modal and bi-modal components. The unification of textual and tabular modalities in MMQA-T5 and PReasM shows that these models achieve major performance improvements through their partial modality integration method. The graph-based SKURG and MGT approaches cannot exceed the performance of these models because structural connections require effective joint representation learning to achieve success.

\begin{table}[ht]
\centering
\caption{Experimental results on MMCoQA.}
\begin{tabular}{lcccc}
\toprule
\multirow{2}{*}{Method} & \multicolumn{2}{c}{Dev} & \multicolumn{2}{c}{Test} \\
\cmidrule(lr){2-3} \cmidrule(lr){4-5}
 & EM & F1 & EM & F1 \\
\midrule
ORConvQA & 1.2 & 3.0 & 1.1 & 1.9 \\
ManymodelQA & 2.3 & 0.7 & 1.8 & 1.0 \\
\midrule
MAE & 19.8 & 26.8 & 22.0 & 28.3 \\
\midrule
Solar & 56.8 & 62.5 & 57.3 & 64.6 \\
\midrule
UniMMQA (T5-Base) & 57.8 & 64.7 & 59.2 & 64.9 \\
UniMMQA (T5-Large) & 62.3 & 69.0 & 63.6 & 70.0 \\
UniMMQA (T5-3B) & \textbf{65.5} & \textbf{71.4} & \textbf{66.7} & \textbf{72.6} \\
\bottomrule
\end{tabular}
\label{tab:mmconvqa}
\end{table}


Solar provides further improvement and achieves competitive results, demonstrating incremental progress toward more effective multimodal understanding. However, its benefits remain restricted. UniMMQA shows superior performance to Solar and all baseline techniques because it uses a unified framework to model all three modalities with rationale. The performance of T5 shows continuous improvement until it reaches T5-3B, which demonstrates how UniMMQA can handle different pre-trained language model sizes. The results demonstrate that multimodal question answering achieves optimal performance through complete modality integration instead of using partial system integration or structural system enhancements.



\begin{table}[ht]
\centering
\caption{Experimental results on ManymodalQA.}
\begin{tabular}{lcc}
\toprule
Method & EM & F1  \\
\midrule
ManymodalQA & 39.7 & - \\
ManymodelQA (w/ type) & 46.3 & - \\
\midrule
UniMMQA (T5-Base) & 45.4 & 45.7 \\
UniMMQA (T5-Large) & 50.0 & 50.4 \\
UniMMQA (T5-3B) & \textbf{52.1} & \textbf{52.4} \\
\bottomrule
\end{tabular}
\label{tab:manymqa}
\end{table}

\begin{table*}[t]
\centering
\caption{Performance comparison of UniMMQA across datasets and modalities (Image, Table, Text).}
\begin{tabular}{llcccccccc}
\toprule
Dataset & Model & \multicolumn{2}{c}{Image} & \multicolumn{2}{c}{Table} & \multicolumn{2}{c}{Text} & \multicolumn{2}{c}{Overall} \\
\cmidrule(lr){3-4} \cmidrule(lr){5-6} \cmidrule(lr){7-8} \cmidrule(lr){9-10}
 &  & EM & F1 & EM & F1 & EM & F1 & EM & F1 \\
\midrule
\multirow{3}{*}{MultimodalQA}
& UniMMQA (T5-Base)  & 66.7 & 69.2 & 65.5 & 71.4 & \textbf{68.3} & \textbf{76.4} & 67.9 & 74.0 \\
& UniMMQA (T5-Large) & 69.8 & 72.2 & 69.9 & 75.6 & \textbf{71.9} & \textbf{80.0} & 71.3 & 77.1 \\
& UniMMQA (T5-3B)    & 73.8 & 76.7 & 73.4 & 79.7 & \textbf{76.2} & \textbf{84.1} & 75.5 & 81.7 \\
\midrule
\multirow{3}{*}{ManymodalQA}
& UniMMQA (T5-Base)  & 46.6 & 46.9 & \textbf{60.7} & \textbf{61.1} & 30.2 & 30.4 & 45.4 & 45.7 \\
& UniMMQA (T5-Large) & 48.5 & 48.6 & \textbf{67.5} & \textbf{68.2} & 34.9 & 35.1 & 50.0 & 50.4 \\
& UniMMQA (T5-3B)    & 49.8 & 50.2 & \textbf{58.0} & \textbf{58.3} & 40.9 & 41.3 & 52.1 & 52.4 \\
\midrule
\multirow{3}{*}{MMConvQA}
& UniMMQA (T5-Base)  & \textbf{73.2} & 75.5 & 33.5 & 40.5 & 66.8 & \textbf{76.3} & 57.8 & 64.7 \\
& UniMMQA (T5-Large) & 73.2 & 75.4 & 38.9 & 46.8 & \textbf{73.3} & \textbf{81.7} & 62.3 & 69.0 \\
& UniMMQA (T5-3B)    & 73.3 & 75.5 & 41.9 & 47.9 & \textbf{77.8} & \textbf{85.1} & 65.8 & 71.4 \\
\bottomrule
\end{tabular}
\label{tab:unimmqa_results}
\end{table*}

\textbf{MMCoQA:} Table \ref{tab:mmconvqa} shows the performance results of MAE, Solar, and UniMMQA against baseline methods on the MMCoQA test-dev set. The baseline models ORConvQA and ManymodelQA show extremely low EM and F1 scores, which demonstrate their inability to perform multimodal conversational question answering tasks that require multiple information sources to be integrated. Their poor performance highlights the inherent difficulty of the task when multimodal reasoning is not effectively modeled.

The introduction of MAE brings a substantial improvement over these baselines, with a large jump in both EM and F1 scores. The system demonstrates its operational success through its modular pipeline, which enables conversational comprehension and adaptive extraction processes. The existing system shows improved performance through its advancements yet still falls behind unified systems which perform at higher levels.


Solar further advances performance significantly, achieving strong results on both Development and Test sets. The study shows that better multimodal integration together with RAG increases answer quality. Nevertheless, UniMMQA maintains superior performance over Solar in all testing conditions. The smallest UniMMQA (T5-Base) version achieves better results than Solar while the larger T5-Large and T5-3B versions provide additional performance improvements.


The continuous performance boost that results from using larger model sizes demonstrates that UniMMQA successfully utilizes larger PLMs to enhance its multimodal reasoning capabilities. The results show that both EM and F1 metrics experienced improvement because the system achieved better exact answer matching and improved semantic correctness. The results show a clear development path that starts from basic weak baselines and progresses to modular systems that use MAE and then reaches complete unified systems which include UniMMQA because it delivers the best performance with scalable results.



\textbf{ManymodalQA:} The performance comparison for the ManymodalQA dataset is demonstrated in Table \ref{tab:manymqa}, which shows how UniMMQA performs against baseline methods. The baseline model, ManymodalQA, shows reasonable performance by leveraging heuristic strategies, while its variant with question type information (ManymodalQA w/ type) achieves a noticeable improvement. The process of selecting which modality to use for answering queries becomes crucial because explicit type information enables better access to appropriate uni-modal QA components.


On the other hand, UniMMQA utilizes unified framework instead of using explicit question type classification. Despite this, it achieves competitive and generally superior performance compared to the standard ManymodalQA baseline. The system demonstrates improved EM and F1 score performance as its model size progresses from T5-Base to T5-3B because it can better understand cross-modal interactions while using more extensive pre-trained language models.


However, UniMMQA (T5-Base) fails to exceed the performance of ManymodalQA w/ type because explicit modality supervision maintains its usefulness in specific situations. The research results demonstrate that heuristic and type-aware methods function as effective baseline methods, while UniMMQA unified modeling system delivers dependable performance which can expand its capabilities without needing extra help to identify different modes of operation.



\subsection{Modality-wise Performance}
Model performance across questions that require information from different modalities is summarized in Table \ref{tab:unimmqa_results}. The UniMMQA system performs better with text-based questions in the MultimodalQA dataset than with image and table-based questions despite its strong overall performance. The model proves better at processing textual information because its visual and tabular data processing capabilities remain restricted. The information loss occurs because images and tables must convert into textual formats which eliminate essential elements that enable accurate reasoning. The process of uniting different modal elements into one reasoning system presents difficulties that need to be solved.


On the ManymodalQA dataset, a different pattern emerges. The study shows that table-based questions produce better results than image-based and text-based questions because their performance exceeds the results shown in MultimodalQA for each specific technology. The ManymodalQA tables use simple table formats because they show only few rows of data, which makes table information easier to understand. The related passages present lengthy and complex content, which creates obstacles for people to understand context and perform reasoning tasks. The model obtains greater advantages from using structured tabular data than from using unstructured text data in this particular situation.


On the other hand, performance on table-based questions in MMConvQA lead to a major drop in performance, while image and text-based questions produce better results. The process of conversational reasoning for tabular data presents multiple challenges because it requires users to handle both table data and evolving dialogue information during their attempt to extract and combine structured content.


\textbf{Overall:} Overall, the results demonstrate that UniMMQA performance depends on the complexity and structure of its underlying modality. The model performs better with textual inputs yet its performance differs across datasets because of varying success in linearizing and combining different modalities. The results demonstrate that unified QA systems require better structural and visual semantics preservation methods and improved multimodal fusion techniques.

\section{Conclusion}
\label{sec:conclusion}
This paper presents a comparative analysis of MAE, Solar, and UniMMQA, highlighting the evolution of multimodal question answering from modular pipelines to unified text-centric frameworks. MAE establishes a strong foundation through explicit modality-aware retrieval and extraction, but remains limited by its reliance on single-modality decisions. Solar advances this direction by transforming heterogeneous inputs into a unified textual space, enabling joint reasoning through a retrieve–rank–generate pipeline, although it suffers from information loss and error propagation. UniMMQA further improves upon this paradigm by incorporating richer modality-to-text transformations and explicit rationale generation, resulting in more robust and scalable performance across benchmarks. Overall, the findings demonstrate that unified approaches leveraging pre-trained language models are more effective for multimodal reasoning. Nevertheless, challenges such as preserving fine-grained multimodal information, mitigating hallucination, and improving computational efficiency remain open, providing important directions for future research.



%



\section*{Acknowledgment}
The author acknowledges the use of ChatGPT (OpenAI) as an assistive tool for language refinement, and conceptual structuring. All generated content was critically reviewed, validated, and appropriately adapted by the author, who takes full responsibility for the accuracy and integrity of the work.



\bibliographystyle{./IEEEtran}
\bibliography{./IEEEexample}

\end{document}